\documentclass{article}

\PassOptionsToPackage{numbers, compress}{natbib}

\usepackage[final]{neurips_2021}

\usepackage[utf8]{inputenc} 
\usepackage[T1]{fontenc}    
\usepackage{hyperref}       
\usepackage{url}            
\usepackage{booktabs}       
\usepackage{amsfonts}       
\usepackage{nicefrac}       
\usepackage{microtype}      
\usepackage{xcolor}         

\usepackage{amsmath,amssymb,amsfonts}
\usepackage{algpseudocode}
\usepackage{graphicx}
\usepackage{subcaption}
\usepackage{textcomp}
\usepackage{xcolor}
\usepackage{hyperref}
\usepackage{url}

\title{Reward-Based Environment States for Robot Manipulation Policy Learning}

\author{
\IEEEauthorblockN{C\'{e}d\'{e}rick Mouliets}
\IEEEauthorblockA{
\textit{Engineering school UPSSITECH}\\
\textit{University of Toulouse III, France} \\
\texttt{cederick.mouliets@univ-tlse3.fr}}
\and
\IEEEauthorblockN{Isabelle Ferran\'{e}}
\IEEEauthorblockA{
\textit{IRIT, Paul Sabatier University}\\
\textit{CNRS, Toulouse, France} \\
\texttt{isabelle.ferrane@irit.fr}}
\and
\IEEEauthorblockN{Heriberto Cuay\'{a}huitl}
\IEEEauthorblockA{
\textit{School of Computer Science} \\
\textit{University of Lincoln, UK}\\
\texttt{hcuayahuitl@lincoln.ac.uk}}
}

\author{%
  C\'{e}d\'{e}rick Mouliets\\
  Engineering school UPSSITECH\\
  University of Toulouse III, France\\
  \texttt{cederick.mouliets@univ-tlse3.fr} \\
  \And
  Isabelle Ferran\'{e} \\
  IRIT, Paul Sabatier University\\
  CNRS, Toulouse, France\\
  \texttt{isabelle.ferrane@irit.fr} \\
  \AND
  Heriberto Cuay\'{a}huitl \\
  School of Computer Science\\
  University of Lincoln, UK\\
  \texttt{hcuayahuitl@lincoln.ac.uk} \\
}

\begin{document}

\maketitle

\begin{abstract}
Training robot manipulation policies is a challenging and open problem in robotics and artificial intelligence. In this paper we  propose a novel and compact state representation based on the rewards predicted from an image-based task success classifier. 
Our experiments---using the Pepper robot in simulation with two deep reinforcement learning algorithms on a grab-and-lift task---reveal that our proposed state representation can achieve up to 97\% task success using our best policies.
\end{abstract}


\section{Introduction}
Teaching a robot new skills easily, quickly and reliably is one of the main goals of robotics and AI. In that way, robots could be of great assistance at home (for example carrying out cleaning or cooking tasks), in factories (carrying out assembly or quality check tasks during production of products), or in public spaces (for instance serving food/drinks or playing games), among other scenarios \cite{AkalinL21,Cuayahuitl2020,KroemerNK21}. Some of the requirements for such an ambitious scientific and engineering goal include understanding robot commands from a human instructor, understanding of the environment and task, acquisition and processing of visual (and audio) perceptions, characterisation of the learning task, ability to select actions for achieving the task, and ability to measure performance, among others. Those requirements and the fact that the world situations are always different, make the problem of skill acquisition of a tall order. The fields of Deep Learning and Reinforcement Learning have been suggested as a promising framework to enable the requirements above. This implies the creation of an agent with a concrete set of states, actions, rewards, and a simulator or logged data. Training a robot means finding the policy that maximises the long-term reward signal via trial-and-error search \cite{LiuNZMD21,IJRR_survey2021}. 

\begin{figure}[ht]
    \centering
    \includegraphics[scale=0.28]{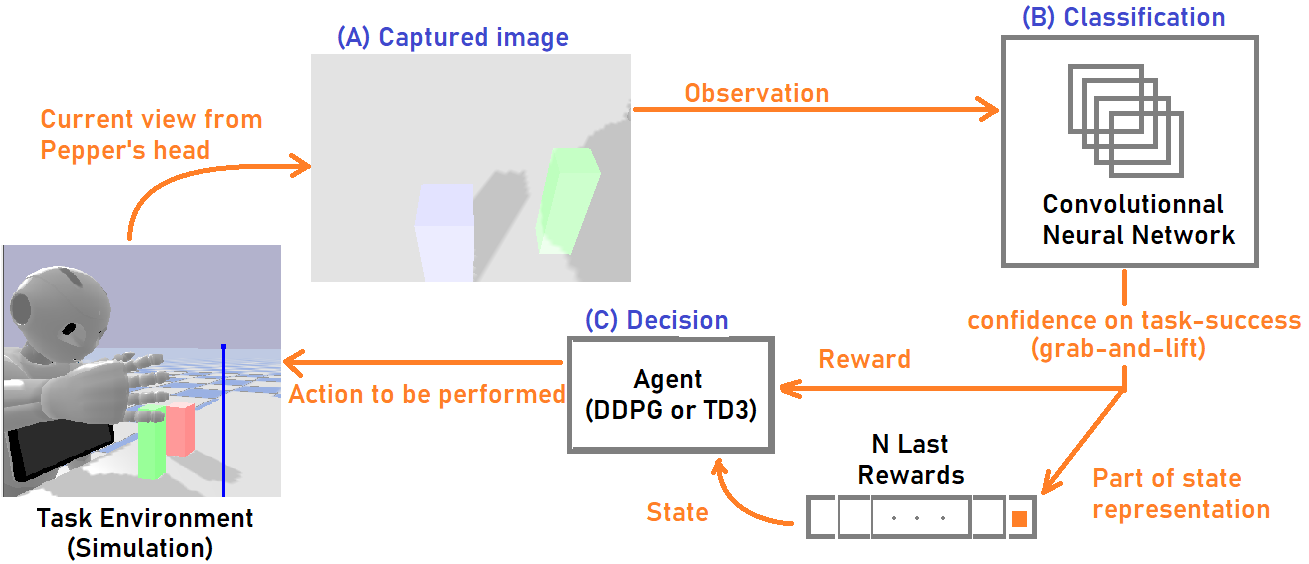}
    \caption{Cycle of the proposed learning agent}
    \label{table_agent}
\end{figure}

The learning scenario of focus in this paper is illustrated in Figure~\ref{table_agent}, where the Humanoid robot Pepper has to carry out a manipulation task in a simulated environment. \textcolor{black}{We focused particularly on a classic task in robotics: grab-and lift. The robot has to grab an object and lift it without dropping it by using reinforcement learning.} The novel aspect in this architecture is the use of compact environment states. In contrast to previous works that use states with learnt features from visual images \cite{SinghYFL19,Mohtasib2021_taros} (not as compact as ours), we use the last $N$=15 rewards to represent our proposed states. 
To our knowledge, this approach has not been studied before. 
Intuitively, those reward-based states can be seen as describing a sequence of task success scores---where  the scores grow as the task  progresses from a time step to the next.

\section{Related Work}
Visual reward learning has been considered as a promising solution for equipping robots with learnable manipulation abilities \cite{SinghYFL19,Mohtasib2021_taros,RL_multitask}. 
This approach measures the success of experiences via reward in robotic tasks by estimating the probability of success from realistic input images  \cite{visual_planning}. While \cite{SinghYFL19} focus on standard Convolutional Neural Networks (CNNs), \cite{Mohtasib2021_taros} extend CNN-based architectures with additional paths on the neural network 
to add more information for better predictions and therefore calculating visual rewards more reliably. These works tested DRL (Deep Reinforcement Learning) agents in classic tasks using a variety of reinforcement algorithms such as DDPG, TD3 or SAC. To do so, different dense and sparse rewards have been investigated, where dense rewards have been found as a more promising alternative than sparse rewards.



Previous works in DRL for robot manipulation such as \cite{FinnTDDLA15,LillicrapHPHETS15,FujimotoHM18,SinghYFL19} represent environment states mostly based on image-based features, where raw images are digested by CNNs in order to learn compact representations that can be used for policy learning. But the idea of using rewards as environment states or part the environment states has not really been studied.  
This highlights the need for a comparison of different state representations, which we identify as a research gap in this paper. In our case, we focus our study on comparing image-based against reward-based state representations. 

\section{Research Methods and Results}
\subsection{Methodology}
We study the overall problem of how to train robot manipulation policies using reinforcement learning in a simulated environment. Given a state-of-the-art algorithm and a defined set of continuous actions, we adopted a reward function based on the success classifier approach as in \cite{SinghYFL19,Mohtasib2021_taros}. To study our proposed state representation, we relied on the following methodology. 
\begin{enumerate}
    \item Create an environment for the grab-and-lift task. 
    \item Create a dataset of success and non-success images.
    \item Train a probabilistic task success classifier (to be used as a reward function).
    \item Define reward-based states and action representations. 
    \item Determine the policy learning algorithm with support for continuous actions.
    \item Train policies using the states, actions and rewards specified above.
    \item Test the learnt policies using  evaluation metrics such as avg. reward and avg. task success.
\end{enumerate}

We used the qiBullet robot simulation environment  \cite{qibullet}, which is powered by the Bullet physics engine and the PyBullet real-time physics simulation framework. To create our simulation environment, we used Unified Robot Description Format  (URDF) files for creating cube and table objects as illustrated in Figure~\ref{table_agent}---the task environment component. \textcolor{black}{To make the robot interact with the cube objects, we specified friction, dampening and stiffness coefficients in the URDF files and put low weights (100g) so the cubes can be lifted. The images for visual perceptions are generated by the 2D camera on the mouth of the robot with resolution $K_{QVGA}$=(320,240).}

\subsection{Visual Reward Classifier and Policy Learning}
\label{reward_classifier}
Instead of manually engineering a reward function, we use learnt rewards from visual observations. To do so, we employ an image-based binary classifier to predict task success (CNN-based). The goal is to generalise situations on the simulation environment in a data efficient manner. 
\textcolor{black}{From the front camera, the robot only sees two cubes on a table.} `Success' images show the robot's hand with the cube grabbed, from different angles, in front of the camera, while `Non-Success' images show the cube outside the robot's hand. 
We randomised the arm positions (on or away from the cube) and generated a total of 6400 labelled images. \textcolor{black}{This dataset of images was created from 10 balanced capture sessions, each session consisting of 200 success images and 440 non-success ones. From one capture session to another, we modified the robot's right arm to introduce variability in its work space. Non-success images are more present because they are more likely to be encountered.} The neural architecture of our reward predictor is based on Tensorflow's image classifier \footnote{\url{https://www.tensorflow.org/tutorials/images/classification}}. 



To learn manipulation policies, we experimented two actor-critic algorithms for decision-making with continuous actions: DDPG \cite{LillicrapHPHETS15}, and TD3  \cite{FujimotoHM18}. 
We used the Stable-Baselines3 framework\footnote{\url{https://github.com/DLR-RM/stable-baselines3}} with default hyperparameters to implement our policies and experimented different state representations. Regarding actions and since the proposed policies only use the right arm of the robot, we focused on four joints: {\it RShoulderPitch, RElbowRoll, RHand, and RShoulderRoll}. Those actions are generated by rotating at a constant speed the angle of each corresponding joint between 0° and 180°. 
\textcolor{black}{A typical learning episode from Figure~\ref{table_agent} is as follows. (A) The robot captures an image of the current situation from the simulated environment. (B) This image is sent to the pre-trained classifier which outputs a confidence score from its last Softmax layer. This score will be considered both as the reward as well as a part of the state representation. (C) Finally and via a deep reinforcement learning algorithm, the agent will choose an action that combines the four joints previously mentioned.}

\subsection{Reward Classification and Manipulation Results}
We split our dataset of success/non-success images (6400 in 10 demos) into three subsets: 8 demos for training, 1 demo for validation and 1 demo for testing. We experimented different configurations in terms of number of epochs and batch size. Our results report that all of our 6 classifiers achieve high performance in terms of classification accuracy (98\% average accuracy on test data-set) and we selected 
our best 
model for predicting visual rewards. \textcolor{black}{Each classifier provides a confidence score of the success of the task which implies that the closer the arm is to the cube, the greater the confidence. Each classifier has been trained and tested using the same dataset split.} The high performance is presumably due to the much simpler simulation environment than the real world. 
The classifier prediction time per image varied between 5.2 and 5.6 milliseconds\footnote{PC specs: 32Gb RAM, Intel Core I7-8650U CPU}.





\begin{table}[b!]
\caption{Best policy learning results}
\begin{center}
\begin{tabular}{l|c|c}
    \hline
    Reinforcement Agent & DDPG | TD3 & DDPG | TD3 \\
    \hline
    State Representation & Avg. Reward & Avg. Task Success (\%) \\
    \hline \hline
    Pixels 320X240 & 2.000 | 2.100 & 76.60 | 78.11 \\
    Pixels 160x120 & {\bf 2.500} | 1.800 & 80.09 | 78.34\\
    Pixels 80x60 & 1.600 | {\bf 2.500} & 28.40 | 88.04 \\
    Image embeddings & 0.003 | 0.001 & 2.7$e^{-5}$ | 1.8$e^{-4}$ \\
    PCA-based images & 1.900 | 1.000 & 11.33 | 57.80 \\
    Last 15 rewards & 1.834 | 2.326 & {\bf 97.04} | {\bf96.78}\\
    \hline
\end{tabular}
\end{center}
\label{ddpg_results}
\end{table}

We trained DDPG policies with six different state representations and compared against their corresponding TD3 policies. \textcolor{black}{These state representations include three different image resolutions (320x240, 160x120, 80x60), Imagenet-based pre-trained embeddings\footnote{\url{https://github.com/jaredwinick/img2vec-keras}} and PCA-based (Principal Component Analysis) images with 50 components \cite{PedregosaVGMTGBPWDVPCBPD11}. Our goal was to improve them using our proposed compact representation based on the last 15 visual rewards (predicted from 160x120 images). PCA and image embeddings enable the significant reduction of amount of information contained in an image. 
While with PCA we reduced the amount of features to 50 components, with image embeddings we obtained vectors of 2048 dimensions.}

\begin{figure}[ht]
    \centering
    \includegraphics[scale=0.58]{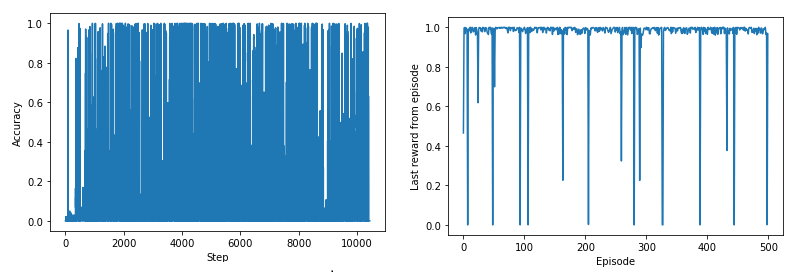}
    \caption{Training and testing of a TD3 policy with 160$\times$120 images, y-axis: the higher the better}
    \label{table_test_train_td3}
\end{figure}

\begin{figure}[h]
    \centering
    \begin{subfigure}[b]{0.157\textwidth}
            \includegraphics[width=\textwidth]{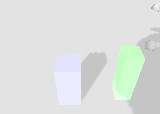}\caption{r1=0.0001}
        \end{subfigure}
   \begin{subfigure}[b]{0.157\textwidth}
            \includegraphics[width=\textwidth]{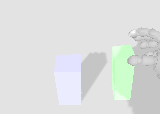}\caption{r2=0.0023} 
        \end{subfigure}
    \begin{subfigure}[b]{0.157\textwidth}
            \includegraphics[width=\textwidth]{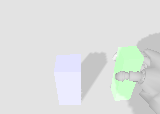}\caption{r3=0.3486}
        \end{subfigure}
   \begin{subfigure}[b]{0.157\textwidth}
            \includegraphics[width=\textwidth]{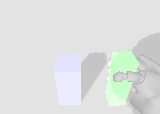}\caption{r4=0.9897}
        \end{subfigure}
    \caption{Example 4-step episode from initial to goal state, [captions]=probabilistic success rewards}
    \label{4_steps_example}
\end{figure}

\begin{figure}[h]
    \centering
    \begin{subfigure}[b]{0.157\textwidth}
            \includegraphics[width=\textwidth]{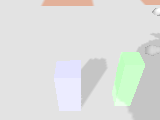}\caption{r1=0.0001}
        \end{subfigure}
   \begin{subfigure}[b]{0.157\textwidth}
            \includegraphics[width=\textwidth]{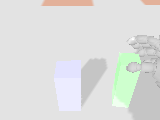}\caption{r2=0.0144}
        \end{subfigure}
    \begin{subfigure}[b]{0.157\textwidth}
            \includegraphics[width=\textwidth]{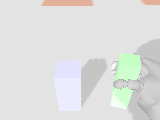}\caption{r3=0.8827}
        \end{subfigure}
   \begin{subfigure}[b]{0.157\textwidth}
            \includegraphics[width=\textwidth]{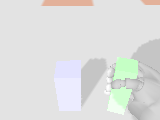}\caption{r4=0.9463}
        \end{subfigure}
    \begin{subfigure}[b]{0.157\textwidth}
            \includegraphics[width=\textwidth]{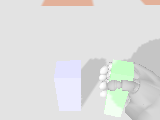}\caption{r5=0.6153}
        \end{subfigure}
   \begin{subfigure}[b]{0.157\textwidth}
            \includegraphics[width=\textwidth]{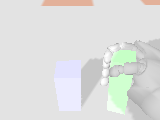}\caption{r6=0.9928}
        \end{subfigure}
    \caption{Example 6-step episode from initial to goal state, [captions]=probabilistic success rewards}
    \label{6_steps_example}
\end{figure}

Our learnt policies take on average 4:45 hrs to train and Table~\ref{ddpg_results} shows their performance out of five runs. While at first sight it looks like TD3 outperforms DDPG in terms of average rewards, that is not the case in terms of average task success. The best DDPG policy obtained a task success of $97.04\%$ and the best TD3 policy obtained $96.78\%$ of task success \textcolor{black}{as shown in Figure~\ref{table_test_train_td3}}. Thus, both learning algorithms can achieve similar performance---but their best policy must be found from multiple runs. \textcolor{black}{The policies were trained and tested for 10000 time steps.
If the robot could not grab the cube after 50 actions, 
we considered the task as a failure and started over another episode as shown in Fig~\ref{table_test_train_td3}. 
} 
Manual inspection showed that when an agent finishes episodes sooner than later, it obtains less rewards. Therefore,
Average Task Success is a more reliable metric in our case. Our best policy uses DDPG with states including the last 15 rewards---a reward-based representation and the most successful in our comparative study. See   episode examples in Figures~\ref{4_steps_example} and~\ref{6_steps_example} with 
successive steps.

\section{Discussion and Concluding Remarks}
This paper proposes a novel methodology for learning manipulation policies based on states represented with sequences of observed rewards derived from a probabilistic task success classifier. Experimental results using simulated interactions in a grab-and-lift task show that our proposed state representation can be useful to learn policies with high-task success (up to 97\%). 
Although other representations need to be investigated further such as tailored image embeddings or pre-trained embeddings with adaptation, our experiments show that it is indeed  possible to achieve high performance with reward-based states instead of image-based states. Combining image-based and reward-based representations is an interesting strand of work that can be explored in the future. 
Two challenges that remain to be investigated with priority are how to efficiently create datasets for training task success classifiers, and how to safely deploy manipulation policies (learnt in simulation) in the real world.

\bibliographystyle{abbrv}
\bibliography{references}  





\end{document}